# Do LLMs Reliably Identify Correct Information Units in Aphasic Discourse?

Jason M Pittman[1]Yesenia Medina-Santos[1], Anton Phillips Jr.[1], Brielle C. Stark[1,2]

Indiana University Bloomington
[1] Department of Speech, Language and Hearing Sciences
[2] Program in Neuroscience

Corresponding Author: Brielle C. Stark, bcstark@iu.edu, 812-855-7760, 2631 E Discovery Parkway, Bloomington IN 47408

## Abstract

**Purpose:** Correct Information Units (CIUs) are central to discourse assessment in aphasia because they quantify communicative informativeness rather than linguistic form alone. However, CIU scoring is time intensive and requires trained raters. This study examined whether instruction-tuned large language models (LLMs) can reliably perform token-level CIU classification from aphasic discourse transcripts.

**Method:** Sixteen picture-description transcripts elicited with the Cat Rescue stimulus were annotated for CIU status according to Nicholas and Brookshire (1993). The sample spanned four severity strata: control, mild, moderate, and severe aphasia. Four publicly available instruction-tuned LLMs were benchmarked under zero-shot and two few-shot prompting conditions across five stratified random seeds. Performance was evaluated against consensus human labels using accuracy, precision, recall, F1, and Cohen's kappa.

**Results:** Zero-shot prompting was insufficient across models. In contrast, few-shot prompting yielded substantial gains and produced competitive performance for three viable models. Mean few-shot F1 scores ranged from 0.776 to 0.817 across Llama-3.1-8B, Qwen2.5-7B, and Mistral-7B, with no significant differences between fixed global and per-chunk local example selection. Phi-3-mini was unstable and did not yield reliable performance. Viable models showed high recall but lower precision, suggesting systematic over-classification of tokens as CIUs. Performance also varied by discourse severity, with the weakest results in more severe aphasia.

**Conclusion:** Few-shot LLM prompting can support automated CIU identification without gradient-based task training, but agreement with human annotation remains insufficient for fully autonomous use. These findings support LLM-based CIU scoring as a promising human-in-the-loop component of clinician-forward discourse assessment systems.

## Introduction

Adults with aphasia often show their greatest communication limitations in discourse (Armstrong, 2000; Doyle et al., 1995; Kintz, 2023; Linnik et al., 2016), where the clinical question is not only how language is formed but whether it successfully conveys intended meaning. Existing natural language processing (NLP) approaches to discourse are powerful for quantifying linguistic form, including lexical features (e.g., lexical diversity, word frequency) and morphosyntactic features (e.g., mean length of utterance, syntactic complexity, part-of-speech distributions). However, these measures primarily characterize how discourse is structured, not whether it successfully conveys the relevant information required by a communicative task. Quantifying informativeness in connected speech is central to aphasia research and clinical assessment (Armes & Richardson, 2020; Boyle et al., 2022; Brisebois et al., 2020; Doyle et al., 1995; Leaman & Edmonds, 2019; Stark & Dalton, 2024; Webster & Morris, 2019). Among the most widely used approaches is the Correct Information Unit (CIU) framework introduced by Nicholas & Brookshire (1993), which scores discourse at the word level according to whether each word is intelligible, accurate, relevant, and informative with respect to the eliciting stimulus or topic. In picture description tasks, CIU analysis provides a principled way to distinguish speech that is merely fluent from speech that successfully conveys content. Because it captures communicative informativeness—and, when normalized to time, communicative efficiency—rather than form alone, CIU scoring has become an important outcome measure of treatment studies and used to assess discourse ability (Boyle et al., 2022; Deng et al., 2024; Leaman & Edmonds, 2019; Stark et al., 2023). In a large international survey study, CIU analysis was done most often in aphasia research centers (~63%) and intensive aphasia programs (~57% in both research and clinical programs). CIUs were used less often in other settings, but still used more often clinically than in research in both the aphasia center/aphasia community groups (~22% clinical vs. ~14% research) and clinical inpatient/outpatient/home settings (~25% clinical vs. ~5% research) (Kiran et al., 2018).

At the same time, CIU analysis is labor intensive. Reliable scoring requires trained raters to inspect transcripts token by token and apply nuanced judgments about relevance, referential accuracy, repetitions, fillers, and other non-informative material. This burden limits scalability in both research and clinical contexts. Large datasets are costly to annotate, repeated longitudinal sampling is difficult to sustain, and routine clinical deployment is constrained by time and training requirements (Cavanaugh et al., 2021). For example, in our dataset of 23 persons with aphasia and 24 persons without aphasia (who had data at two timepoints: test and retest), it took two trained raters approximately five months to hand annotate and crosscheck about 6 samples x 47 people x 2 timepoints, and then do reliability scoring for ~20% of that data (Stark et al., 2023). CIU specificity notwithstanding, discourse analysis has had limited uptake clinically because of barriers related to time, tools, and resources (Bryant et al., 2017; Cruice et al., 2020; Stark et al., 2021). These practical barriers to discourse analysis have motivated longstanding interest in automation (Liu et al., 2023). Recent work has shown that supervised machine learning methods can predict CIU labels from transcribed discourse with encouraging performance (Ng et al., 2025), suggesting that at least some aspects of the construct are computationally tractable. However, supervised approaches require labeled training data drawn from the same task domain, and their generalizability depends on the availability of annotated corpora that remain scarce in aphasia research and also limit generalizability to new tasks or corpora.

The emergence of instruction-tuned large language models (LLMs) raises a different possibility. Rather than learning CIU classification solely from task-specific labeled data, an LLM might be able to apply the formal CIU definition directly from natural-language instructions, or infer the operational decision procedure from a small number of worked examples. This is especially compelling for CIU scoring because the task is conceptually structured yet judgment-intensive. To classify a token as a CIU, a rater must integrate local lexical content, surrounding context, discourse pragmatics, and knowledge of the eliciting stimulus. These are precisely the kinds of mixed linguistic and semantic inferences that contemporary LLMs are designed to perform. If successful, such models could offer a low-data route to automated discourse analysis that is more accessible than conventional supervised pipelines.

Whether LLMs can do this reliably remains an open empirical question. On one hand, the CIU framework is explicit enough to be described procedurally: Nicholas and Brookshire's criteria define the construct clearly (their original 1993 appendix has an exhaustive appendix, though there are some missing elements, e.g., it does not explicitly discuss sound effects, such as onomatopoeias), and the scoring task can be presented as a constrained token-level labeling problem. On the other hand, applying CIU criteria consistently is not trivial. CIU judgments often depend on subtle distinctions between informative and non-informative words, on whether a word contributes meaning only in combination with surrounding tokens, and on whether an utterance accurately characterizes a particular picture rather than merely sounding plausible. These challenges may be amplified by aphasia severity. Speech produced by individuals with more severe impairment is more likely to contain neologisms (made-up words), paraphasias (word errors), agrammatism or paragrammatism (a lack of or incorrect grammar), phonological approximations, circumlocutions, fragments, false starts, and reduced local or global coherence, all of which can make it harder to determine whether a given token is intelligible, relevant, and informative in context. In such cases, CIU scoring may depend not only on the formal rules themselves but also on the rater's ability to interpret atypical or degraded output in context.

Relatedly, CIU judgments are unlikely to be entirely independent of listener experience. Raters with substantial exposure to aphasic speech may be better able to infer intended referents, recognize recurrent error patterns, or recover communicative intent from partial or distorted forms, whereas less experienced raters may classify the same tokens more conservatively as unintelligible or non-informative. Further, it is most likely that the person scoring CIUs (if hand scoring) is the treating clinician, who is likely to have some familiarity with the speaking pattern of the person with aphasia, and this is unlikely to ever be the case with an LLM. This raises an important measurement question: reliability may vary not only as a function of the scoring rubric, but also as a function of the interaction between transcript difficulty and rater expertise. The problem for LLM evaluation, then, is not simply whether a model can "understand" CIUs in principle, but whether it can apply the construct reproducibly across levels of discourse impairment and in a manner that approximates trained human judgment. Moreover, because LLMs are known to vary in instruction-following fidelity, structured output compliance, and sensitivity to prompt design, any claim that they can automate CIU scoring must address both overall performance and robustness under the very conditions that make human scoring difficult.

This question matters because CIUs are more than a convenient annotation target. Prior work, has shown that CIU-based measures have meaningful psychometric properties in aphasia, supporting their validity as indicators of informativeness and communicative success (Brookshire & Nicholas, 1994; Stark et al., 2023). That literature strengthens the case for automation: if CIUs index clinically important aspects of discourse, then a scalable method for deriving them could substantially expand the reach of discourse-based assessment. Automated CIU scoring could facilitate larger observational studies, improve reproducibility across laboratories, and enable more feasible monitoring of discourse outcomes in intervention research. It could also help bridge the gap between rich discourse analysis and practical deployment in clinical settings.

The present study therefore asks a focused question: can publicly available instruction-tuned LLMs reliably identify CIUs in aphasic discourse transcripts from a standard picture-description task? We compare two broad routes to task acquisition. In one, models are instructed directly in the standardized CIU procedure derived from Nicholas and Brookshire. In the other, models are additionally exposed to previously labeled examples, allowing in-context learning to approximate concept induction from prior analyses. We evaluate four open or publicly accessible instruction-tuned models under zero-shot and few-shot prompting conditions on token-level CIU classification in transcripts elicited by the Cat Rescue picture-description stimulus.

Our aim is not to claim that LLMs replace human discourse analysts, nor to treat any single benchmark as definitive. Rather, we test whether modern small-to-mid-sized LLMs can operationalize a clinically meaningful discourse construct without gradient-based training on the target task. Demonstrating such capacity would have both theoretical and practical significance: theoretically, it would suggest that CIU scoring is representable as a promptable linguistic decision procedure; practically, it would identify a promising path toward lightweight, reproducible, and data-efficient automation of discourse analysis in aphasia.

## Methods

### *Dataset and Annotation*

Speech transcripts were drawn from an existing dataset of picture-description samples collected using the Cat Rescue stimulus (Nicholas & Brookshire, 1993), in which participants describe a line-drawing depicting a cat stranded in a tree, a man climbing to rescue it, a dog barking below, and a fire truck arriving in the background. The dataset comprised 16 transcripts spanning four severity strata: control (no communication impairment), mild, moderate, and severe aphasia. This dataset is a subset of the corpus used in Pittman et al. (2025), who evaluated a zoo of supervised machine learning classifiers on automated CIU scoring from the same stimulus using 130 transcripts. The present study uses a smaller subset of 16 transcripts to evaluate whether large language models can achieve competitive performance without requiring labeled training data.

CIU annotation followed the criteria established by Nicholas and Brookshire (1993). Under these criteria, a Correct Information Unit is defined as any intelligible, informative word that is relevant to the picture being described and accurately characterises it. Tokens were classified as non-CIU if they were unintelligible, irrelevant to the stimulus, inaccurate, or represented fillers, false starts, or repetitions. Each token in all 16 transcripts was independently annotated by two trained raters. Inter-rater reliability was assessed using Pearson's r, yielding $r = 0.985$, indicating near-perfect agreement on CIU counts across transcripts. Disagreements were resolved by consensus prior to analysis.

Following annotation, the dataset comprised 1,893 tokens across 16 transcripts. Of these, 1,191 tokens (65.5% of all words) were classified as CIU. Schema validation confirmed that all required fields were present for each token record, and an invariant check confirmed that no token with word_label = 1 (CIU) had an associated word_label = 0 entry, ensuring annotation consistency. The class imbalance (approximately 2:1 CIU to non-CIU) is characteristic of picture-description samples from speakers without severe aphasia and was preserved in all downstream splits and evaluations.

### *Experimental Design*

The study employed a fully crossed design of four models by three prompting modes, replicated across five random seeds, yielding 60 experimental cells in total. The 16 transcripts were partitioned into a prompt split and an evaluation split using stratified random sampling to preserve representation across the four severity strata. Five transcripts were allocated to the prompt split, which provided the pool from which few-shot examples were drawn; the remaining eleven transcripts formed the evaluation set on which all classification metrics were computed. Stratified splitting was performed independently at each of the five seeds (2025–2029), producing seed-specific evaluation sets of approximately 985–1,155 tokens depending on the split.

Zero-shot conditions leveraged greedy decoding and are therefore deterministic across seeds. The five replications of each zero-shot condition thus represent a single deterministic result and are reported accordingly; standard deviations for zero-shot conditions are zero by construction and should not be interpreted as reflecting sampling variability.

### *Language Models*

Four publicly available instruction-tuned large language models were evaluated. All models were accessed via HuggingFace Hub and loaded in their instruction-tuned variants to ensure compatibility with chat-style prompting. Table 1 summarises the models, their HuggingFace identifiers, parameter counts, licenses, and access requirements.

**Table 1.** Models evaluated in the benchmarking study. All models are instruction-tuned variants loaded via HuggingFace Hub.

| Model | HuggingFace ID | Params | License | Access |
| --- | --- | --- | --- | --- |
| Llama-3.1-8B-Instruct | meta-llama/Llama-3.1-8B-Instruct | 8B | Llama 3.1 Community | Gated (Meta) |
| Qwen2.5-7B-Instruct | Qwen/Qwen2.5-7B-Instruct | 7B | Apache 2.0 | Open |
| Mistral-7B-Instruct-v0.3 | mistralai/Mistral-7B-Instruct-v0.3 | 7B | Apache 2.0 | Open |
| Phi-3-mini-4k-instruct | microsoft/Phi-3-mini-4k-instruct | 3.8B | MIT | Open |

Note. Llama-3.1-8B-Instruct requires acceptance of Meta's Llama 3.1 Community License Agreement via HuggingFace gated access. All other models are freely available under open licenses.

Model selection was motivated by practical deployment considerations. The 7–8 billion parameter range represents the largest class of models that can be run on a single consumer GPU or institutional compute node without multi-GPU parallelism, making results directly reproducible in typical clinical research computing environments. Phi-3-mini at 3.8 billion parameters was included to assess whether a smaller, more resource-efficient model could achieve comparable performance, with the expectation that its reduced capacity might limit instruction-following reliability on a structured labeling task.

### *Prompt Design*

*Structure*

All prompts were structured using a two-part system and user message format applied via each model's native chat template (tokenizer.apply_chat_template()). This ensured that model-specific special tokens required for instruction-following were correctly applied across all four architectures. The system prompt provided the task definition, CIU criteria, picture context, and output format constraint. The user message contained the target tokens formatted as a structured table.

The system prompt included an explicit description of the Cat Rescue picture stimulus to provide the semantic grounding necessary for CIU classification. Specifically, that the picture depicts a cat stuck in a tree, a man climbing to rescue it, a dog barking below, and a fire truck arriving in the background. This context was included because CIU classification requires determining whether a token accurately and relevantly describes the specific stimulus, not whether it is informative in the abstract. Each inference chunk was presented to the model as a two-column token table formatted with INDEX and TOKEN headers and enclosed in XML-style <input> tags. Each row contained a zero-indexed token position and its normalised surface form. This tabular format was chosen over a plain numbered list to reduce positional ambiguity when models produce output arrays, and to provide a visual structure that mirrors the expected JSON output format. A worked example covering all major decision categories including fillers, unintelligible tokens, function words, content CIU tokens, and repetitions was included in the system prompt enclosed in <example_input> and <example_output> tags. The output format constraint "YOUR RESPONSE MUST START WITH [ AND CONTAIN ONLY THE JSON ARRAY" was included in both the system prompt and the user message.

*Chunking*

Because transformer models have finite context windows and because long token sequences produce proportionally long output arrays that risk exceeding the max_new_tokens budget, transcripts were segmented into chunks of at most 16 tokens prior to inference. A minimum chunk size of 8 tokens was enforced: remainder chunks shorter than 8 tokens were merged into the preceding chunk rather than submitted as a standalone inference call. This prevents degenerate single- or double-token calls that would produce unreliable classifications and waste allocation budget. The chunk size of 16 was selected to ensure that the model's chain-of-thought preamble (approximately 580 tokens) plus the JSON output array (approximately 350 tokens for 16 labels) fit comfortably within max_new_tokens = 3000.

*Modes*

Three prompting modes were evaluated. Table 2 defines each mode and its distinguishing characteristics.

**Table 2.** Prompting modes evaluated in the benchmarking study.

| Mode | Description | Example (n) | Scope |
|---|---|---|---|
| **z_shot_local** | No worked examples provided. System prompt includes task description and output format constraint only. Model must infer CIU criteria from the prompt definition. | 0 | N/A |
| **few_shot_local** | Worked examples drawn from the prompt split (n=5 transcripts). For each inference chunk, examples are selected from local prompt pool and included in the user message preceding the target tokens. | Variable | Per-chunk local |
| **few_shot_global** | A fixed set of worked examples constructed from the prompt split and held constant across all inference chunks and all seeds. Examples cover all major CIU decision categories. | Fixed global | Global constant |

Note. All three modes used identical system prompts and token table formatting. Modes differ only in whether and how worked examples are incorporated into the user message.

***Inference Configuration***

All inference was conducted using the HuggingFace Transformers pipeline API. Table 3 summarises the key configuration parameters, their values, and the rationale for each choice.

**Table 3.** Inference configuration parameters applied uniformly across all model-mode conditions.

| Parameter | Value | Rationale |
|---|---|---|
| **Decoding strategy** | Greedy (do_sample=False) | Ensures deterministic output for reproducibility; avoids probability tensor instability at near-zero temperatures on MPS/CPU. |
| **max_new_tokens** | 3000 | Accommodates chain-of-thought preamble (~580 tokens) plus JSON output array (~350 tokens) per chunk. |
| **chunk_size** | 16 tokens | Limits per-chunk token count to fit within max_new_tokens budget with headroom; prevents output truncation. |
| **min_chunk_size** | 8 tokens | Merges short remainder chunks into predecessor to avoid degenerate single-token inference calls. |
| **dtype** | float16 (CUDA) / float32 (CPU/MPS) | float16 halves GPU memory footprint; float32 used on CPU/MPS where float16 is unsupported or causes conversion errors. |
| **Chat template** | tokenizer.apply_chat_template() | Applies model-specific special tokens (e.g., <\|eot_id\|>, [INST]) required for instruction-following behaviour. |
| **Random seeds** | 2025, 2026, 2027, 2028, 2029 | Five seeds used for stratified dataset splitting; zero-shot conditions are deterministic and seed-invariant. |
| **Hardware** | IU Quartz cluster, NVIDIA GPU nodes | Jobs submitted via SLURM to gpu partition; each model run on a single GPU node (--gres=gpu:1). |

Output parsing used a bracket-scan extraction strategy: the parser scanned model output for the first [ character and extracted the substring through the matching ] to recover the JSON array regardless of any chain-of-thought preamble preceding it. This approach was adopted after validation confirmed that all three viable models consistently produced reasoning text prior to the JSON array, even when instructed to output only the array. The bracket-scan strategy achieved 100% parse recovery on the validation transcript set.

***Supervised Baseline***

A supervised machine learning baseline was trained and evaluated on the same data split to provide a reference comparison for the LLM results. The baseline employed a Linear Support Vector Classifier (LinearSVC) with a feature set comprising TF-IDF weighted unigram-to-trigram token n-grams concatenated with one-hot encoded severity stratum indicators. The TF-IDF representation captures local lexical patterns predictive of CIU status, while the severity feature encodes the global communication profile of each speaker. The pipeline was implemented using scikit-learn and serialised for reproducibility.

The baseline was trained on 924 tokens from the prompt split and evaluated on 231 tokens from the evaluation split, using the same stratified partition as the LLM evaluation. This design ensures that baseline and LLM metrics are computed on identical held-out token sets and are therefore directly comparable. The baseline achieved accuracy = 0.710, precision = 0.749, recall = 0.851, F1 = 0.796, and Cohen's kappa = 0.300 on the evaluation set.

It is important to note that the baseline was trained on labeled data from the same dataset, while the LLMs received no training examples — only in-context demonstrations via the few-shot prompt. This comparison is therefore not controlled for the amount of supervision provided. The baseline result is reported as a reference point to contextualise LLM performance relative to a conventional supervised approach, not as a directly equivalent experimental condition. For comparison, Pittman et al. (2025) reported a best CIU classification accuracy of 0.824 using k-nearest neighbours trained on 130 transcripts from the same stimulus task, though differences in dataset size and feature engineering preclude direct performance comparisons between that work and the present study.

***Evaluation Metrics and Statistical Testing***

Classification performance was assessed at the token level using five metrics: accuracy, precision, recall, F1 score, and Cohen's kappa. Accuracy reflects the overall proportion of correctly classified tokens. Precision and recall are reported with respect to the CIU class (label = 1), reflecting the model's ability to correctly identify CIU tokens without over- or under-prediction. F1 is the harmonic mean of precision and recall and is the primary comparative metric given the approximately 2:1 class imbalance in the evaluation set. Cohen's kappa quantifies agreement between model predictions and human reference labels beyond chance and provides a more conservative index of classification quality than accuracy or F1 under class imbalance.

Confidence intervals for F1 and kappa were estimated using stratified bootstrap resampling with 1,000 iterations per cell, producing 95% CIs reported in the per-run summary. All metrics were computed independently for each of the five seeds and summarised as means and standard deviations across seeds in the aggregated results.

Pairwise statistical comparisons between model-mode conditions were conducted using McNemar's chi-square test on pooled token-level predictions across all five seeds. McNemar's test is appropriate for paired binary classification comparisons where the same token set is evaluated under two conditions and the test statistic is based on the number of discordant predictions (n01 and n10). Pooling across seeds increases statistical power by aggregating evidence from multiple evaluation sets rather than treating each seed independently. The significance threshold was set at $\alpha = .05$. Given the large number of pairwise comparisons performed ($k = 55$ total; $k = 15$ for viable few-shot conditions), Bonferroni-corrected thresholds are reported alongside uncorrected p-values where relevant.

Parse quality was assessed for each cell by recording the number of output chunks that yielded a parseable JSON array of the expected length. Cells with a parse success rate below 90% were flagged for review prior to metric computation. No viable model cell fell below this threshold.

Severity-stratified analyses were conducted by partitioning the evaluation tokens by the stratum label of the transcript from which they were drawn (control, mild, moderate, severe) and computing all metrics separately within each stratum. The severe stratum comprised only 11 tokens across the dataset and is reported descriptively only; no inferential statistics are reported for this stratum.

### *Reproducibility*

All experimental code, configuration files, and SLURM job scripts are versioned and available in the study repository (Pittman, 2026). The pipeline is structured as a sequence of discrete, checkpointed stages: data preparation, stratified splitting, LLM inference, output parsing, and metric computation. A completed-cell checkpoint file records which model-mode-seed combinations have been successfully processed, enabling interrupted runs to resume without reprocessing completed cells. Configuration parameters governing inference, chunking, and splitting are centralised in a single config.yaml file, and all random seeds are explicitly parameterised. Prompt templates are stored in a separate ciu_prompts.yaml file and rendered at runtime using parameter substitution to ensure reproducibility of the exact prompt text used for each cell.

All LLM inference was conducted on Indiana University's Quartz research computing cluster using NVIDIA GPU nodes allocated via SLURM. Models were loaded in their exact HuggingFace Hub versions as identified by the model strings in Table 1. Float16 precision was used on CUDA devices; float32 was used on CPU fallback. The baseline classifier was trained and serialised using scikit-learn and can be reloaded from the distributed joblib file for independent verification.

## Results

This study evaluated four instruction-tuned large language models: Llama-3.1-8B-Instruct, Qwen2.5-7B-Instruct, Mistral-7B-Instruct-v0.3, and Phi-3-mini-4k-instruct. We tested each model on the task of automated token-level CIU classification from picture-description speech transcripts. Each model was evaluated under three prompting conditions: zero-shot (*z_shot_local*), few-shot with per-chunk examples drawn from the local prompt set (*few_shot_local*), and few-shot with a fixed global example set held constant across all chunks (*few_shot_global*). All conditions were replicated across five random seeds (2025–2029). Performance was assessed using accuracy, precision, recall, F1, and Cohen's kappa against human-annotated reference labels. McNemar's chi-square tests were used for pairwise significance testing across pooled predictions. All reported metrics reflect means and standard deviations across the five seeds unless otherwise noted.

A total of 60 experimental cells were completed (4 models × 3 modes × 5 seeds). One model, Phi-3-mini, exhibited severe instability across seeds and is treated separately from the three viable models. Note that zero-shot conditions used greedy decoding (do_sample=False), rendering predictions deterministic across seeds; accordingly, zero-shot standard deviations are zero and those five cells represent a single deterministic condition.

### *Overall Model and Prompting Performance*

Table 4 presents accuracy, precision, recall, F1, and Cohen's kappa for all model-mode combinations averaged across five seeds. Figure 1 plots mean F1 with standard deviation error bars for all four models across all three prompting modes. Figure 2 shows all four performance metrics for the three viable models in the two few-shot conditions.

Table 4. Classification performance (mean and SD across 5 seeds) for all model-mode conditions. Phi-3-mini values are greyed to indicate non-viability. F1 values are bolded for viable models.

| Model | Mode | Accuracy | Precision | Recall | F1 | Cohen's κ |
|---|---|---|---|---|---|---|
| Llama-3.1-8B | Zero-shot | 0.581 (0.000) | 0.784 (0.000) | 0.513 (0.000) | 0.620 (0.000) | 0.196 (0.000) |
| | Few-shot (local) | 0.707 (0.027) | 0.715 (0.023) | 0.938 (0.024) | 0.811 (0.016) | 0.207 (0.086) |
| | Few-shot (global) | 0.711 (0.010) | 0.714 (0.009) | 0.955 (0.015) | 0.817 (0.006) | 0.189 (0.045) |
| Qwen2.5-7B | Zero-shot | 0.472 (0.000) | 0.864 (0.000) | 0.248 (0.000) | 0.385 (0.000) | 0.125 (0.000) |
| | Few-shot (local) | 0.651 (0.021) | 0.752 (0.017) | 0.708 (0.066) | 0.728 (0.028) | 0.241 (0.026) |
| | Few-shot (global) | 0.694 (0.021) | 0.740 (0.016) | 0.832 (0.056) | 0.782 (0.021) | 0.270 (0.049) |
| Mistral-7B | Zero-shot | 0.537 (0.000) | 0.739 (0.000) | 0.437 (0.000) | 0.549 (0.000) | 0.134 (0.000) |
| | Few-shot (local) | 0.688 (0.040) | 0.709 (0.034) | 0.893 (0.071) | 0.789 (0.032) | 0.216 (0.091) |
| | Few-shot (global) | 0.677 (0.031) | 0.703 (0.025) | 0.867 (0.055) | 0.776 (0.023) | 0.220 (0.084) |
| Phi-3-mini | Few-shot (local) | 0.375 (0.083) | 0.742 (0.150) | 0.116 (0.215) | 0.145 (0.240) | 0.011 (0.024) |
| | Few-shot (global) | 0.407 (0.090) | 0.642 (0.379) | 0.197 (0.267) | 0.242 (0.267) | 0.022 (0.021) |

Note. Values in parentheses are standard deviations. Zero-shot SD = 0.000 reflects deterministic greedy decoding.

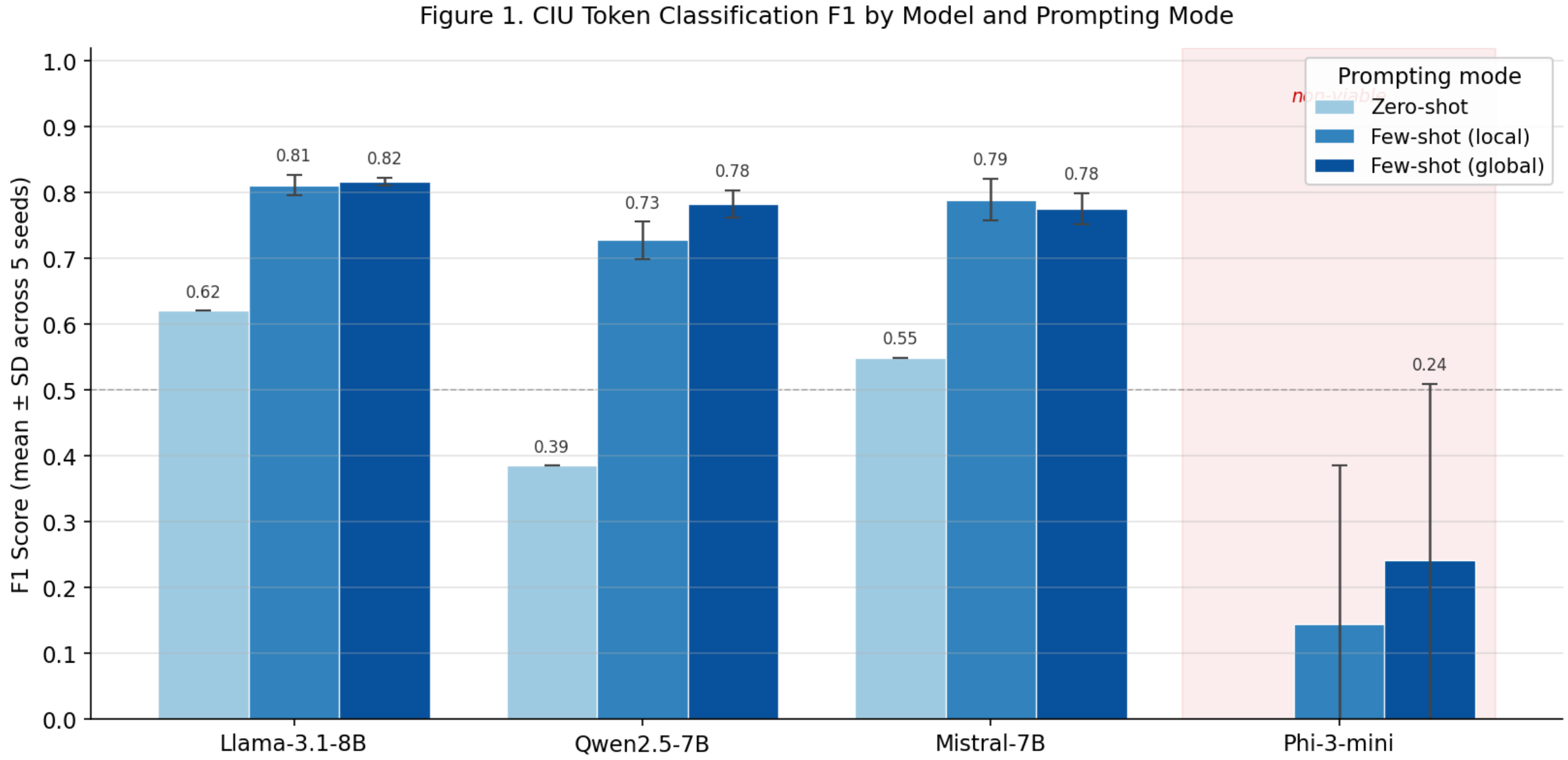

Figure 1. Mean F1 score (± SD across 5 seeds) for all four models across three prompting modes. Error bars represent one standard deviation. The shaded red column indicates Phi-3-mini, which exhibited non-viable performance across all conditions. The dashed horizontal line marks F1 = 0.50.

Among the three viable models, few-shot prompting produced substantial and consistent improvements over zero-shot prompting. Llama-3.1-8B achieved the highest mean F1 under both few-shot conditions: F1 = 0.817 (SD = 0.006) under few_shot_global and F1 = 0.811 (SD = 0.016) under few_shot_local. Qwen2.5-7B followed with F1 = 0.782 (SD = 0.021) under few_shot_global and F1 = 0.728 (SD = 0.028) under few_shot_local. Mistral-7B produced F1 = 0.776 (SD = 0.023) under few_shot_global and F1 = 0.789 (SD = 0.032) under few_shot_local, placing it narrowly ahead of Qwen2.5-7B in the local condition. Zero-shot performance was markedly lower across all three viable models (Llama: F1 = 0.620; Qwen: F1 = 0.385; Mistral: F1 = 0.549), confirming that in-context examples are necessary for reliable token-level CIU classification.

All three viable models showed consistently high recall under few-shot conditions (range: 0.708–0.955), indicating a tendency to over-predict CIU status — that is, to classify non-CIU tokens as CIU — rather than under-predict. This asymmetry is consistent with a model bias toward marking tokens as communicatively informative in the absence of strong negative evidence.

Cohen's kappa values across viable few-shot conditions ranged from 0.189 to 0.270, reflecting fair-to-moderate agreement with human annotation beyond chance. The relatively modest kappa values relative to the higher F1 scores reflect the class imbalance in the evaluation set (approximately 65.5% CIU tokens), which inflates accuracy- and recall-based metrics. Kappa provides a more conservative and appropriate index of agreement for imbalanced binary classification.

Figure 2. Performance Metrics for Viable Models — Few-Shot Conditions

Few-shot (global)

Few-shot (local)

Figure 2. Accuracy, precision, recall, and F1 (mean ± SD) for the three viable models under few-shot global (left) and few-shot local (right) prompting conditions.

### *Effect of Prompting Mode*

Zero-shot prompting was uniformly insufficient for reliable CIU classification. All three viable models showed large and statistically significant drops in F1 relative to both few-shot conditions (all McNemar $p < .001$; see Table 2). Llama-3.1-8B declined from a few-shot global mean of F1 = 0.817 to a zero-shot F1 = 0.620; Qwen2.5-7B from 0.782 to 0.385; and Mistral-7B from 0.776 to 0.549. These results indicate that providing worked examples of the CIU labeling task in the prompt is a prerequisite for acceptable automated scoring performance.

Within the few-shot conditions, the difference between few_shot_global and few_shot_local was negligible for all three viable models. McNemar tests comparing the two few-shot modes within each model were all non-significant (Llama: $\chi^2(1) = 0.25$, $p = .617$; Mistral: $\chi^2(1) = 0.017$, $p = .897$; Qwen: $\chi^2(1) = 0.004$, $p = .947$). This finding has practical significance: the computationally simpler approach of maintaining a fixed global example set is functionally equivalent to the more complex per-chunk local selection strategy.

### *Pairwise Statistical Comparison*

Table 5 presents McNemar chi-square tests for all pairwise comparisons among the three viable models in few-shot conditions. Conditions labeled 'global' and 'local' correspond to few_shot_global and few_shot_local respectively. $n_{01}$ counts tokens misclassified by Condition A but correctly classified by Condition B; $n_{10}$ counts the reverse.

With the exception of two borderline comparisons, the three viable models were statistically indistinguishable from one another in few-shot conditions. Qwen2.5-7B (few_shot_global) was significantly outperformed by Llama-3.1-8B (few_shot_global) at $\chi^2(1) = 4.86$, $p = .027$, with $n_{10} = 144 > n_{01} = 108$ indicating Llama correctly classified more discordant tokens. A parallel comparison between Qwen2.5-7B (few_shot_global) and Llama-3.1-8B (few_shot_local) similarly reached significance ($p = .050$). However, neither comparison survives Bonferroni correction for 15 simultaneous tests (α_corrected = .003), and all remaining 13 pairwise comparisons were non-significant (all $p > .06$).

Table 5. McNemar pairwise significance tests for viable model few-shot conditions. Significant comparisons ($\alpha$ = .05) are indicated in bold red. $n_{01}$ and $n_{10}$ reflect the number of discordant token-level predictions pooled across 5 seeds.

| Condition A | Condition B | $n_{01}$ | $n_{10}$ | $\chi^2$ | p | Sig. ($\alpha$=.05) |
|---|---|---|---|---|---|---|
| Qwen2.5-7B (global) | Qwen2.5-7B (local) | 114 | 114 | 0.004 | 0.947 | No |
| Qwen2.5-7B (global) | Llama-3.1-8B (global) | 108 | 144 | 4.861 | 0.027 | **Yes*** |
| Qwen2.5-7B (global) | Llama-3.1-8B (local) | 94 | 124 | 3.858 | 0.050 | **Yes*** |
| Qwen2.5-7B (global) | Mistral-7B (global) | 164 | 168 | 0.027 | 0.869 | No |
| Qwen2.5-7B (global) | Mistral-7B (local) | 167 | 174 | 0.106 | 0.745 | No |
| Qwen2.5-7B (local) | Llama-3.1-8B (global) | 192 | 228 | 2.917 | 0.088 | No |
| Qwen2.5-7B (local) | Llama-3.1-8B (local) | 167 | 197 | 2.310 | 0.129 | No |
| Qwen2.5-7B (local) | Mistral-7B (global) | 197 | 201 | 0.023 | 0.881 | No |
| Qwen2.5-7B (local) | Mistral-7B (local) | 221 | 228 | 0.080 | 0.777 | No |
| Llama-3.1-8B (global) | Llama-3.1-8B (local) | 53 | 47 | 0.250 | 0.617 | No |
| Llama-3.1-8B (global) | Mistral-7B (global) | 153 | 121 | 3.507 | 0.061 | No |
| Llama-3.1-8B (global) | Mistral-7B (local) | 130 | 101 | 3.394 | 0.065 | No |
| Llama-3.1-8B (local) | Mistral-7B (global) | 154 | 128 | 2.216 | 0.137 | No |
| Llama-3.1-8B (local) | Mistral-7B (local) | 141 | 118 | 1.869 | 0.172 | No |
| Mistral-7B (global) | Mistral-7B (local) | 117 | 120 | 0.017 | 0.897 | No |

* $p < .05$. Note that two comparisons reaching significance (Qwen global vs. Llama global, p = .027; Qwen global vs. Llama local, p = .050) would not survive Bonferroni correction for 15 simultaneous comparisons ($\alpha$_corrected = .003) and should be interpreted with caution.

These results suggest that among 7-to-8 billion parameter instruction-tuned models, the choice of model architecture is less consequential than the choice to use few-shot prompting. Once in-context examples are provided, Llama-3.1-8B, Qwen2.5-7B, and Mistral-7B converge to broadly equivalent classification performance on this task.

**Phi-3-mini: Non-Viable Performance**

Phi-3-mini-4k-instruct failed to produce reliable CIU classifications under any prompting condition. Mean F1 across few-shot conditions was 0.242 (SD = 0.267) for few_shot_global and 0.145 (SD = 0.240) for

few_shot_local, with standard deviations exceeding or approaching the mean in both cases. Figure 3 illustrates the per-seed instability: seed 2025 produced F1 = 0.662 under few_shot_global, while seeds 2026 and 2027 collapsed to F1 = 0.019 and 0.000 respectively, and seeds 2028–2029 converged at F1 = 0.264. This pattern is inconsistent with random variation and reflects structural instruction-following failures that differ across random seeds.

The near-zero recall values observed across most Phi-3-mini cells (e.g., recall = 0.116, SD = 0.215 for few_shot_local) indicate that the model predominantly failed to identify CIU tokens rather than misclassifying non-CIU tokens, a pattern consistent with the model defaulting to a near-null output strategy rather than engaging with the classification prompt. Phi-3-mini's 3.8 billion parameter count likely places it below the capability threshold required for reliable token-level labeling of this complexity at zero- and few-shot.

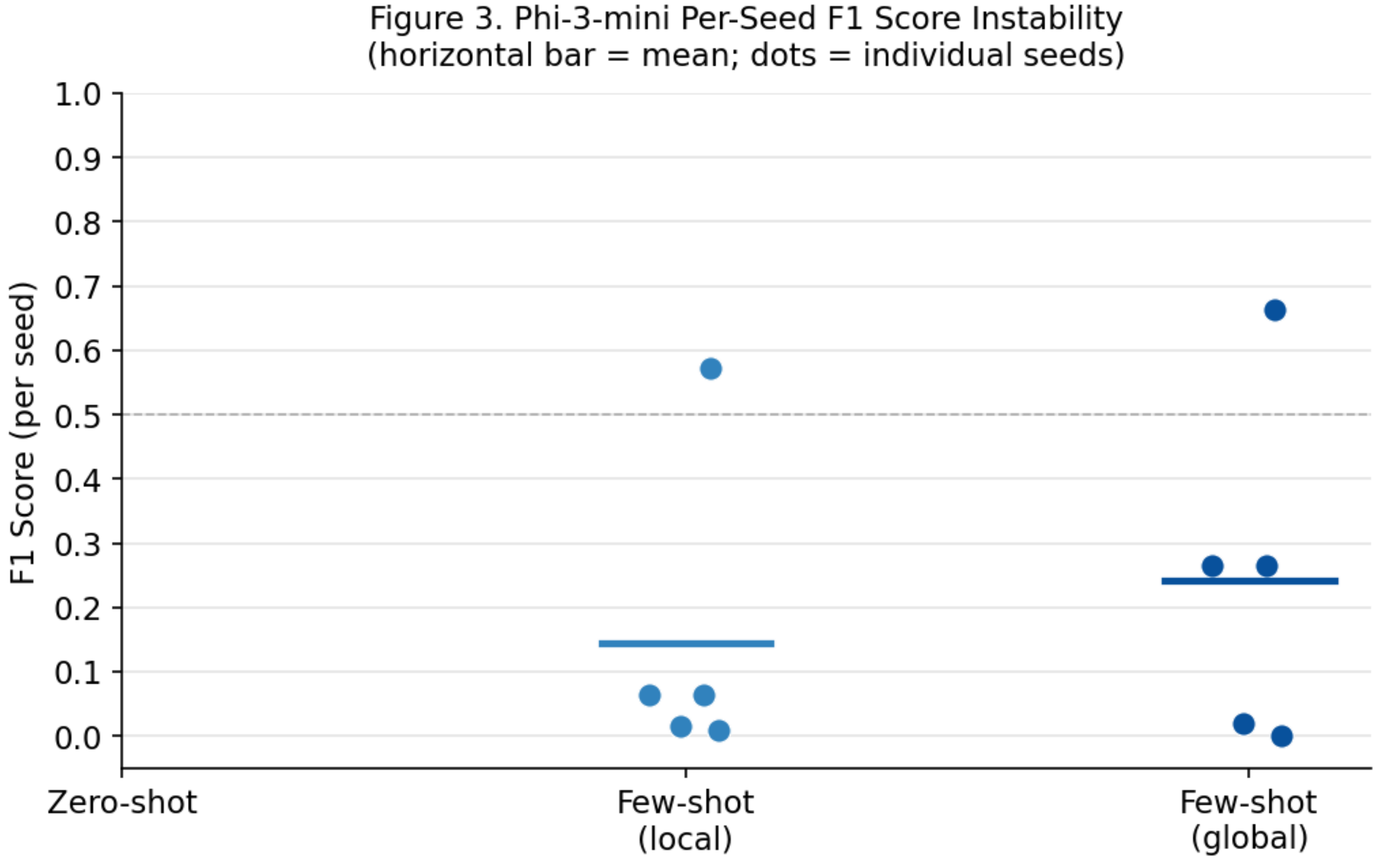

Figure 3. Phi-3-mini per-seed F1 score by prompting mode. Each point represents one seed (n = 5); horizontal bars indicate the mean. The near-zero results for most seeds and the extreme variance across seeds indicate non-viable performance across all conditions.

***Performance by Severity Strata***

Figure 4 presents mean F1 by severity stratum for the three viable models under few_shot_global, averaged across the five seeds. Performance was markedly uneven across severity levels, with mild severity consistently producing the strongest results across all models.

For mild severity transcripts, all three viable models achieved strong performance (F1 range across models and seeds: 0.77–0.90). For control severity transcripts, performance was moderate (F1 range: 0.58–0.88) with higher variance across seeds. Moderate severity transcripts produced the weakest results among the three analyzable strata: all models showed high recall but substantially depressed precision in this stratum, indicating systematic over-prediction of CIU status for speakers with moderate communication impairment. This pattern likely reflects that moderate-severity speakers produce more ambiguous token contexts, where the CIU decision boundary is less clearly cued by the surrounding utterance structure.

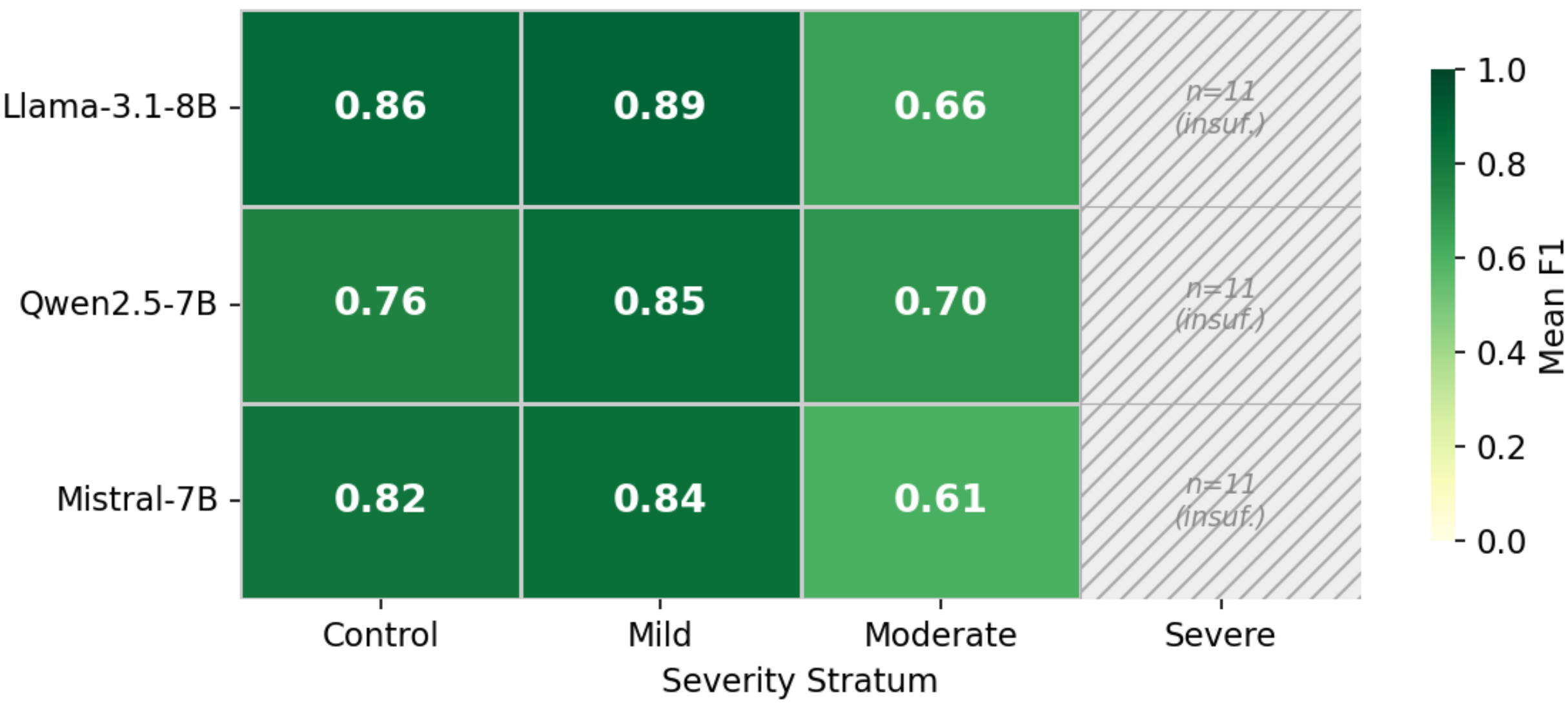

Figure 4. Mean F1 by severity stratum for the three viable models under few-shot global prompting (5-seed average). The severe stratum (hatched) contained only 11 tokens and is reported descriptively only. Color intensity reflects F1 from 0 (white) to 1 (dark green).

The severe severity stratum contained only 11 tokens across the 16-transcript dataset, precluding reliable statistical analysis. F1 values for this stratum were uniformly zero or near-zero across all models and seeds due to extreme class imbalance. These results should not be interpreted substantively. The absence of sufficient severe-stratum data is an acknowledged limitation of the current dataset and a target for future data collection.

***Summary of Findings***

Taken together, four findings characterize the results of this benchmarking study. First, few-shot prompting is a necessary condition for viable automated CIU classification: zero-shot performance was uniformly insufficient across all models. Second, among three viable 7-to-8B parameter models, performance differences were largely non-significant after correction for multiple comparisons, with Llama-3.1-8B showing a marginal uncorrected advantage over Qwen2.5-7B in the global condition. Third, the choice between few-shot global and few-shot local prompting strategies did not significantly affect classification performance, supporting the use of the simpler global approach in applied settings. Fourth, Phi-3-mini at 3.8 billion parameters was not viable for this task, and its results are excluded from primary conclusions. Performance degradation for moderate-severity speakers and the insufficient severe-stratum sample size represent key limitations for clinical translation.

## Discussion

The present study asked whether publicly available instruction-tuned large language models can reliably identify Correct Information Units in aphasic discourse transcripts. The answer is qualified but encouraging. Across three viable 7–8B models, token-level CIU classification was clearly possible, but only when models were given worked examples in context. Zero-shot prompting was uniformly inadequate, whereas few-shot prompting yielded substantial gains in F1 and brought performance into a range that was competitive with a simple supervised baseline trained on labeled data from the same corpus. At the same time, agreement beyond chance remained only fair to moderate by Cohen's kappa, and performance was uneven across severity strata. Taken together, these findings suggest that LLMs can approximate CIU scoring under constrained conditions, but they do not yet support fully autonomous use in clinical decision-making.

One important implication is conceptual. CIU scoring has often been treated as a labor-intensive human judgment task that depends on nuanced interpretation of discourse context, communicative relevance, and stimulus fidelity. The present results suggest that this construct is at least partially representable as a promptable decision procedure. In other words, an LLM can do more than reproduce shallow lexical patterns: with appropriate instructions and examples, it can approximate a clinically meaningful discourse annotation framework that requires integrating local token identity, utterance context, and world knowledge about the pictured scene. That finding is theoretically interesting because it places CIU analysis in a class of discourse judgments that appear accessible to in-context learning rather than requiring task-specific model training.

At the same time, the results also show the limits of that claim. Few-shot prompting was not a marginal optimization; it was the difference between non-viable and viable performance. This suggests that the formal definition of a CIU, although explicit, is not by itself sufficient for stable task execution in current small-to-mid-sized LLMs. Models benefited substantially from exposure to worked examples that concretized difficult distinctions such as fillers versus meaningful function words, repetitions versus informative reiterations, and locally plausible but stimulus-inaccurate tokens. From a practical perspective, this is useful: the simpler few-shot global prompt performed as well as the more complex per-chunk local retrieval approach, indicating that a fixed and carefully curated example bank may be sufficient for applied use.

The error profile is equally important. Across viable few-shot conditions, models tended to show high recall and lower precision, indicating a bias toward over-labeling words as CIUs. Clinically and methodologically, this matters because overestimating informativeness could lead to inflated estimates of communicative success, particularly in speakers whose discourse contains many ambiguous or partially relevant tokens. This pattern aligns with the severity-stratified results, where moderate aphasia yielded the weakest performance among the analyzable strata. Moderate-severity discourse may be especially challenging because it often contains enough semantic content to appear informative while also including distortions, circumlocutions, or incomplete propositions that make CIU boundaries uncertain. In that sense, the moderate group may represent the hardest region of the decision space: not clearly intact, but not so severely impaired that most tokens are obviously non-CIU.

These severity effects also connect to a broader issue raised in the introduction: CIU scoring is not independent of the listener. Human raters with more experience interpreting aphasic discourse may be better able to recover intended meaning from atypical forms, whereas inexperienced raters may apply criteria more conservatively. The present study used consensus human labels as the reference standard, but those labels still reflect judgments situated within a particular training context. LLMs may therefore be learning not only the CIU rubric, but also an implicit style of expert interpretation encoded in the examples. This is not a weakness unique to LLMs; rather, it highlights that automated CIU scoring inherits the same construct-validity questions that apply to human scoring.

The performance of the LLM in the current study should be interpreted in the context of human inter-rater performance. For example, in a recent study by Deng and colleagues, inter-rater reliability was performed on a random subset of 16 discourse samples, with 8 from a mild anomic aphasia group and 8 from the control group. Reliability was calculated using point-to-point agreement (Deng et al., 2024). For CIU counts, inter-rater reliability was 92% in the control group and 90% in the anomic aphasia group. Intra-rater values showed the same pattern: consistently high overall, but modestly lower in the aphasia group (CIU counts: 95% controls vs. 92% aphasia). Even in a relatively mild group—these participants had anomic aphasia, with median AQ = 86.9, preserved fluency, and no dysarthria or apraxia—the aphasia samples yielded slightly lower agreement than neurotypical controls, and neither inter-rater and intra-rater agreement was not 100%. If trained human rater agreement is already somewhat reduced in a mild, relatively fluent group, it is reasonable to expect that automated scoring may become even more challenging as aphasia severity increases and discourse contains more ambiguity, distortions, fragments, circumlocutions, or reduced coherence.

Future work should therefore examine not only model-human agreement, but also how agreement varies as a function of transcript severity, rater expertise, and annotation protocol. In relation to transcript severity, as we have shown above, a model may agree well with human raters on control or mild aphasia samples, where discourse is relatively interpretable, but perform much worse on moderate or severe samples, where there are more fragments, neologisms, approximations, false starts, or context-dependent productions. In relation to rater expertise, not all humans score CIUs in the same way, as discussed above. Experienced aphasia raters may recover intended meaning from distorted productions better than novice raters, and they may be more comfortable distinguishing nonstandard but still informative productions from true non-CIUs. That means "human gold standard" is partly shaped by who the humans are. Future work should therefore test whether model agreement is closer to expert raters than novice raters, and whether the model mirrors expert interpretive tendencies or more conservative novice tendencies. Finally, human agreement also depends on how scoring is done. Are raters working from transcript only, or transcript plus audio? Did they receive formal CIU training? Did they score independently and reconcile disagreement by consensus? Were edge cases such as sound effects, gestures, or unintelligible approximations handled with a written decision rule? Different protocols can change the reference labels themselves. Therefore, future work should ask whether LLM performance is robust across different scoring setups, not only against one particular annotation pipeline.

**Clinical Implications**

The clinical promise of these findings lies in augmentation rather than replacement. Few-shot LLMs could plausibly serve as first-pass annotators that pre-label transcripts, identify likely CIUs, and reduce the time burden on trained clinicians or researchers. Even if human review remains necessary, reducing manual token-by-token scoring could make discourse analysis more feasible in treatment studies, longitudinal monitoring, and larger observational datasets. The fact that competitive performance was achieved with relatively small, publicly available models also improves the feasibility of local deployment in research or health-system environments where privacy, cost, and reproducibility are important.

More broadly, the present benchmarking study should be understood as part of the development pathway for REAL-EVAL ("("Real-world Evaluation of Aphasic Language") Studio rather than as a standalone attempt to automate CIU scoring in isolation. REAL-EVAL is being built by our team as a clinician-forward assessment environment for discourse, with CIUs representing one candidate metric of propositional informativeness within that broader framework. The value of the current results is therefore translational: they suggest that a lightweight LLM-based CIU pipeline may be viable enough to embed in a platform where outputs are editable, patient-adaptive, and interpreted alongside other discourse measures. Seen in this way, the contribution of the present study is not to argue that LLMs should replace expert discourse analysts, but to show that CIU analysis may be operationalized in a form suitable for integration into a clinically usable tool.

### Assumptions and Limitations

Several limitations constrain interpretation. First, the dataset was small and restricted to a single elicitation task, the Cat Rescue picture description. This limits both statistical power and generalizability to other discourse genres such as narrative retell, procedural discourse, or conversation. Second, the severe stratum was too sparse for meaningful inference, preventing strong conclusions about the population in whom automated support may be most needed and where the data is likely to be most sparse (i.e., the group that is likely to produce the fewest tokens). Third, the evaluation was conducted on tokenized transcripts rather than audio, so the models did not face challenges related to speech recognition, prosody, or acoustic intelligibility. Fourth, comparison with the supervised baseline was informative but not strictly controlled, because the baseline had access to labeled training data whereas the LLMs relied only on in-context examples. Finally, prompt design, output parsing, and chunking decisions were highly engineered; performance may therefore reflect the quality of the surrounding pipeline as much as intrinsic model capability.

### Future Work

Several next steps follow directly. The most immediate is scale: larger and more diverse corpora are needed, especially with more moderate and severe aphasia and with multiple discourse tasks. Prompt optimization is another priority, including systematic manipulation of number of examples, example composition, and decision-rule wording. It will also be important to move beyond binary token classification toward transcript-level CIU summaries and clinically interpretable endpoints such as percent CIUs and CIUs per minute. More broadly, future work should test hybrid pipelines in which LLM judgments are combined with conventional NLP features, retrieval of annotated exemplars, or uncertainty estimates that flag low-confidence segments for human review. The most useful endpoint is unlikely to be a fully autonomous scorer, but a robust human-in-the-loop system that makes discourse-level assessment faster, more reproducible, and more clinically practical. Along such lines, future work may also consider frontier models with larger parameter sizes and potentially more robust training to gauge whether more recent LLMs have higher innate capability relative to CIU analysis.

## Conclusions

This study provides initial evidence that instruction-tuned large language models can approximate Correct Information Unit scoring from aphasic discourse transcripts, but only under constrained conditions and not yet at a level that justifies autonomous clinical use. Across three viable 7–8B models, few-shot prompting substantially improved token-level classification relative to zero-shot prompting, bringing performance into a range that was competitive with a simple supervised baseline despite the absence of gradient-based task training. At the same time, agreement beyond chance remained only fair to moderate, error patterns were clinically meaningful, and performance was uneven across severity levels, particularly for more ambiguous discourse. These results therefore support a cautious interpretation: CIU scoring appears to be sufficiently structured to be operationalized by contemporary LLMs, but current implementations remain dependent on prompt design, worked examples, and human oversight.

The broader significance of this work is translational. We do not view automated CIU identification as an end in itself, nor as a replacement for expert discourse analysis. Rather, we view it as one promising component of a larger clinician-forward assessment framework. In that context, the most important contribution of the present study is to show that a lightweight, publicly available LLM pipeline can produce useful first-pass estimates of propositional informativeness that may reduce analytic burden while preserving clinician authority over interpretation. This is directly aligned with the development of REAL-EVAL Studio, in which automated discourse metrics are intended to be transparent, editable, patient-adaptive, and clinically interpretable over time.

Accordingly, the practical endpoint of this line of work is not a fully autonomous CIU scorer, but a robust human-in-the-loop discourse analysis system that makes clinically meaningful measurement more feasible in real workflow. Further validation on larger, more diverse datasets and across broader discourse contexts is necessary, but the present findings suggest that LLM-based CIU scoring is a credible step toward scalable, reproducible, and clinically usable discourse assessment in aphasia.

## Data Availability

CIU data is available via request to Brielle Stark, bcstark@iu.edu.

## Acknowledgments

Original data collection and analysis was funded through the ASHFoundation New Investigator Grant to BC Stark.

## References

Armes, E., & Richardson, J. (2020). *The Relationship between Narrative Informativeness and Psychosocial Outcomes in Chronic Stroke-Induced Aphasia*.

Armstrong, E. (2000). Aphasic discourse analysis: The story so far. *Aphasiology*, *14*(9), 875–892. https://doi.org/10.1080/02687030050127685

Boyle, M., Akers, C. M., Cavanaugh, R., Hula, W. D., Swiderski, A. M., & Elman, R. J. (2022). Changes in discourse informativeness and efficiency following communication-based group treatment for chronic aphasia. *Aphasiology*, *0*(0), 1–35. https://doi.org/10.1080/02687038.2022.2032586

Brisebois, A., Brambati, S. M., Désilets-Barnabé, M., Boucher, J., García, A. O., Rochon, E., Leonard, C., Desautels, A., & Marcotte, K. (2020). The importance of thematic informativeness in narrative discourse recovery in acute post-stroke aphasia. *Aphasiology*, *34*(4), 472–491. https://doi.org/10.1080/02687038.2019.1705661

Brookshire, R. H., & Nicholas, L. E. (1994). Speech sample size and test-retest stability of connected speech measures for adults with aphasia. *Journal of Speech and Hearing Research*, *37*(2), 399–407. https://doi.org/10.1044/jshr.3702.399

Bryant, L., Spencer, E., & Ferguson, A. (2017). Clinical use of linguistic discourse analysis for the assessment of language in aphasia. *Aphasiology*, *31*(10), 1105–1126. https://doi.org/10.1080/02687038.2016.1239013

Cavanaugh, R., Kravetz, C., Jarold, L., Quique, Y., Turner, R., & Evans, W. S. (2021). Is There a Research–Practice Dosage Gap in Aphasia Rehabilitation? *American Journal of Speech-Language Pathology*, *30*(5), 2115–2129. https://doi.org/10.1044/2021_AJSLP-20-00257

Cruice, M., Botting, N., Marshall, J., Boyle, M., Hersh, D., Pritchard, M., & Dipper, L. (2020). UK speech and language therapists ' views and reported practices of discourse analysis in aphasia rehabilitation. *International Journal of Language and Communication Disorders*, 1–26. https://doi.org/10.1111/1460-6984.12528

Deng, B.-M., Liang, L.-S., Zhao, J.-X., Zheng, H.-Q., & Hu, X.-Q. (2024). Correct Information Unit Analysis in Different Discourse Tasks Among Persons With Anomic Aphasia Based on Mandarin AphasiaBank. *American Journal of Speech-Language Pathology*, *33*(2), 800–813. https://doi.org/10.1044/2023_AJSLP-23-00217

Doyle, P. J., Goda, A. J., & Spencer, K. A. (1995). The Communicative Informativeness and Efficiency of Connected Discourse by Adults With Aphasia Under Structured and Conversational Sampling Conditions. *American Journal of Speech-Language Pathology*, *4*(4), 130–134. https://doi.org/10.1044/1058-0360.0404.130

Kintz, S. (2023). Discourse Characteristics in Aphasia. In A. P.-H. Kong (Ed.), *Spoken Discourse Impairments in the Neurogenic Populations: A State-of-the-Art, Contemporary Approach* (pp. 23–36). Springer International Publishing. https://doi.org/10.1007/978-3-031-45190-4_2

Kiran, S., Cherney, L. R., Kagan, A., Haley, K. L., Antonucci, S. M., Schwartz, M., Holland, A. L., & Simmons-Mackie, N. (2018). Aphasia assessments: A survey of clinical and research settings. *Aphasiology*, *32*(sup1), 47–49. https://doi.org/10.1080/02687038.2018.1487923

Leaman, M. C., & Edmonds, L. A. (2019). Revisiting the Correct Information Unit: Measuring Informativeness in Unstructured Conversations in People With Aphasia. *American Journal of Speech-Language Pathology*, *28*(3), 1099–1114. https://doi.org/10.1044/2019_AJSLP-18-0268

Linnik, A., Bastiaanse, R., & Höhle, B. (2016). Discourse production in aphasia: A current review of theoretical and methodological challenges. *Aphasiology*, *30*(7), 765–800. https://doi.org/10.1080/02687038.2015.1113489

Liu, H., MacWhinney, B., Fromm, D., & Lanzi, A. (2023). Automation of Language Sample Analysis. *Journal of Speech, Language, and Hearing Research*, 1–13. https://doi.org/10.1044/2023_JSLHR-22-00642

Ng, S.-I., Ambadi, P. S., Mueller, K. D., Liss, J., & Berisha, V. (2025). *Advancing Automated Spatio-Semantic Analysis in Picture Description Using Language Models* (arXiv:2510.05128). arXiv. https://doi.org/10.48550/arXiv.2510.05128

Nicholas, L. E., & Brookshire, R. H. (1993). A System for Quantifying the Informativeness and Efficiency of the Connected Speech of Adults With Aphasia. *Journal of Speech, Language, and Hearing Research*, *36*(2), 338–350. https://doi.org/10.1044/jshr.3602.338

Pittman, J. (2026). Aphasia-llm-ciu. Zenodo. https://doi.org/10.5281/zenodo.19484477

Stark, B. C., Alexander, J. M., Hittson, A., Doub, A., Igleheart, M., Streander, T., & Jewell, E. (2023). Test-Retest Reliability of Microlinguistic Information Derived From Spoken Discourse in Persons With Chronic Aphasia. *Journal of Speech, Language, and Hearing Research: JSLHR*, *66*(7), 2316–2345. https://doi.org/10.1044/2023_JSLHR-22-00266

Stark, B. C., & Dalton, S. G. (2024). A scoping review of transcription-less practices for analysis of aphasic discourse and implications for future research. *International Journal of Language & Communication Disorders*, *59*(5), 1734–1762. https://doi.org/10.1111/1460-6984.13028

Stark, B. C., Dutta, M., Murray, L. L., Fromm, D., Bryant, L., Harmon, T. G., Ramage, A. E., & Roberts, A. C. (2021). Spoken Discourse Assessment and Analysis in Aphasia: An International Survey of Current Practices. *Journal of Speech, Language, and Hearing Research*, *23 Sept*. https://doi.org/10.1044/2021_JSLHR-20-00708

Webster, J., & Morris, J. (2019). Communicative Informativeness in Aphasia: Investigating the Relationship Between Linguistic and Perceptual Measures. *American Journal of Speech-Language Pathology*, *28*(3), 1115–1127. https://doi.org/10.1044/2019_AJSLP-18-0256